\documentclass[conference]{IEEEtran}
\IEEEoverridecommandlockouts

\usepackage{cite}
\usepackage{mathtools}
\usepackage{amsmath,amssymb,amsfonts}
\usepackage{graphicx}
\usepackage{textcomp}
\usepackage{xcolor}
\usepackage{amsmath}
\usepackage{multirow}
\usepackage{multicol}
\usepackage{caption}
\usepackage{array}
\usepackage{subcaption}
\usepackage{balance}
\usepackage{wrapfig}
\usepackage{hyperref}
\usepackage{multirow}
\usepackage{multicol}
\usepackage{color}
\usepackage{parskip}
\usepackage{colortbl}
\usepackage{booktabs}
\usepackage{flushend}
\usepackage{hyperref}
\usepackage{todonotes}
\usepackage{comment}

\usepackage{cite}

\usepackage[utf8]{inputenc}
\usepackage[english]{babel}

\usepackage{algpseudocode}
\usepackage{algorithm}

\newcolumntype{L}{>{\centering\arraybackslash}m{3cm}}

\makeatletter
\newcommand{\linebreakand}{%
  \end{@IEEEauthorhalign}
  \hfill\mbox{}\par
  \mbox{}\hfill\begin{@IEEEauthorhalign}
}
\makeatother

\def\BibTeX{{\rm B\kern-.05em{\sc i\kern-.025em b}\kern-.08em
    T\kern-.1667em\lower.7ex\hbox{E}\kern-.125emX}}
    
\begin{document}

\title{Predicting Early Indicators of Cognitive Decline from Verbal Utterances\\

\thanks{This work is funded by the National Institutes of Health (NIH) under award number 1K01LM012439-01}
}

\author{\IEEEauthorblockN{Swati Padhee}
\IEEEauthorblockA{\textit{Computer Science and Engineering} \\
\textit{Wright State University}\\
Dayton, OH \\
padhee.2@wright.edu}
\and
\IEEEauthorblockN{Anurag Illendula}
\IEEEauthorblockA{\textit{Computer Science} \\
\textit{New York University, Courant}\\
New York City, NY\\
\textit{ai2153@nyu.edu}\\
 \\
}
\and
\IEEEauthorblockN{Megan Sadler}
\IEEEauthorblockA{\textit{Biological Sciences} \\
\textit{Wright State University}\\
Dayton, OH \\
sadler.13@wright.edu}
\and
\IEEEauthorblockN{Valerie L.Shalin}
\IEEEauthorblockA{\textit{Dept. of Psychology} \\
\textit{Wright State University}\\
Dayton, OH \\
valerie.shalin@wright.edu}
%\and
\linebreakand %
\IEEEauthorblockN{Tanvi Banerjee}
\IEEEauthorblockA{\textit{Computer Science and Engineering} \\
\textit{Wright State University}\\
Dayton, OH \\
tanvi.banerjee@wright.edu}
\and
\IEEEauthorblockN{Krishnaprasad Thirunarayan}
\IEEEauthorblockA{\textit{Computer Science and Engineering} \\
\textit{Wright State University}\\
Dayton, OH \\
t.k.prasad@wright.edu}
\and
\IEEEauthorblockN{William L. Romine}
\IEEEauthorblockA{\textit{Biological Sciences} \\
\textit{Wright State University}\\
Dayton, OH \\
william.romine@wright.edu}
}

\maketitle

\begin{abstract}
Dementia is a group of irreversible, chronic, and progressive neurodegenerative disorders resulting in impaired memory, communication, and thought processes.
In recent years, clinical research advances in brain aging have focused on the earliest clinically detectable stage of incipient dementia, commonly known as mild cognitive impairment (MCI).
Currently, these disorders are diagnosed using a manual analysis of neuropsychological examinations. 
We measure the feasibility of using the linguistic characteristics of verbal utterances elicited during neuropsychological exams of elderly subjects to distinguish between elderly control groups, people with MCI, people diagnosed with possible Alzheimer’s disease (AD) and probable AD. 
We investigated the performance of both theory-driven psycholinguistic features and data-driven contextual language embeddings in identifying different clinically diagnosed groups. 
Our experiments show that a combination of contextual and psycholinguistic features extracted by a Support Vector Machine improved distinguishing the verbal utterances of elderly controls, people with MCI, possible AD, and probable AD. 
This is the first work to identify four clinical diagnosis groups of dementia in a highly imbalanced dataset. Our work shows that machine learning algorithms built on contextual and psycholinguistic features can learn the linguistic biomarkers from verbal utterances and assist clinical diagnosis of different stages and types of dementia, even with limited data. 
\end{abstract}


\section{Introduction}
\label{sec:Introduction}

Dementia manifests as a set of cognitive and social symptoms that interfere with daily functioning in older adults with symptoms ranging from aggression, and depression, to unfocused movement. 
Although Alzheimer’s disease (AD) is the most common type of dementia, there are others such as Vascular Dementia, Dementia with Lewy Bodies (DLB), Mixed Dementia, and Parkinson’s Disease. An intermediate transitional stage between expected cognitive decline due to normal aging and the more-serious decline of dementia is Mild Cognitive Impairment (MCI)~\cite{petersen2004mild}.  
As individuals suffering from MCI show a decline in cognitive skills in their daily lives, it is difficult to determine whether the symptoms result from healthy aging or early onset of dementia. 
Hence, identifying the early indicators of incipient dementia in a pre-clinical stage can help patients and their families prepare for early intervention.

Current laboratory methods for the early detection of dementia are limited to expensive and invasive brain imaging techniques such as computerized tomography (CT) scans, positron emission tomography (PET) scans, magnetic resonance imaging (MRI)~\cite{nensa2014clinical}, neurological examinations, mental status examinations, and physical exams. 
The feasibility of an automatic, cost-effective, and non-invasive solution for early and frequent monitoring of a larger cohort of the aging population depends on identifying dementia at the earliest possible stage of cognitive impairment. Moreover, people suffering from MCI may not exhibit the symptoms of dementia, such as impaired judgment or trouble with reasoning. Recent studies have demonstrated the limitations of neuropsychological examinations for diagnosing dementia. 
Prior studies indicate that linguistic patterns captured in the verbal utterances can be behavioral biomarkers of dementia and MCI because they are affected by neurodegenerative disorders (ND) that control cognitive, speech, and language processes~\cite{reilly2010cognition,verma2012semantic}. According to J.L. Locke  \cite{locke1997theory}, language deficits can affect the lexical and syntactic processes governing language and verbal utterances. A significant body of work has attempted to process text and speech samples to detect cognitive impairment using machine learning (ML) and natural language processing (NLP)~\cite{masrani2017domain,fraser2016linguistic,orimaye2014learning}.

We investigated a combination of data-driven and theory-driven approaches in designing a computational clinical diagnostic model to identify early symptoms of cognitive deficits due to healthy aging (a confounder) from those of mild cognitive impairment (MCI), which can also be a precursor to dementia leading to AD. The potential clinical utility of our work is the ability to predict the MCI phenotype before it develops into AD. 
Our key contributions are as follows:
\begin{itemize}
    \item To the best of our knowledge, this is the first study to formulate a multi-class problem that distinguishes healthy elderly controls, people with MCI, possible AD, and probable AD, using verbal utterances. 
    \item We demonstrate the efficacy of machine learning models built on both \textit{data-driven} contextual language embeddings and \textit{theory-driven} psycholinguistic features to identify different stages and types of dementia from the analysis of verbal utterances.
\end{itemize}

\begin{table*}[htbp]
\centering 
\caption{Example transcripts of Cookie Theft picture description by participants from different diagnosis groups (AD:Alzheimer's Disease).}
 \scalebox{0.8}{
\vline
\begin{tabular}{| m{0.2\columnwidth} |m{2.0\columnwidth}|}
\hline
 \textbf{Diagnosis Group}  &\textbf{Transcript of verbal description of the Cookie Theft Picture}  \\

    \hline
    Healthy elderly control 
    & the scene is in the  in the kitchen.the mother is wiping dishes and the water is running on the floor.a child is trying to get  a boy is trying to get cookies outta  a jar and he's about to tip over on a stool.uh the little girl is reacting to his falling.uh it seems to be summer out.the window is open.the curtains are blowing.it must be a gentle breeze.there's grass outside in the garden.uh mother's finished certain of the  the dishes.kitchen's very tidy. the mother seems to have nothing in the house to eat except cookies in the cookie jar.uh the children look to be almost about the same size.perhaps they're twins.they're dressed for summer warm weather.um you want more ?    the mother's in a short sleeve dress.  I'll have to say it's warm.\\
    \hline
     Mild Cognitive Impairment (MCI)
    &  is it alright to say mother and s son and daughter ?  mother is wiping dishes but she's overflowing the sink bowl .water is running on the floor .outside I can't tell if it's a cloudy day or a great day but you can see some shrubbery and  and bushes . and the son  and the tilting stool is about to fall .taking cookies for he and his sister out of the cookie jar. so somebody's gon gonna  s have some cleanup work to do .and the daughter is   course reaching her left arm and hand and elbow up to get  receive a cookie .but he's gonna  go crashing on down the floor where the water is going, the suds water. \textcolor{blue}{I do the dishes sometimes at home. but with the dishwasher it's just pots and pans.   fry pans, do our eggs in the morning.  I use rubber gloves because my  skin cracks around my nails. `. }  mhm.  the drapes are alright.I don't see anything else . .\\
    \hline
    Possible AD 
    &. the boy's gonna  fall.. she's running the sink over... that kid seem to getting in the cookie jar. so  ...  . and she's standing in water working.    \\            
    \hline
    Probable AD 
    & hm a lady  a lady and her children. children.  the lady is wash  washing dishes.and  ... okay. yeah. . the children have cookies.. \textcolor{blue}{the father isn't coming.father didn't come in yet.}and the lady is getting it ready.get  ... it's a water with that. water went down.it's gonna  fall over. 
     \\
    \hline
\end{tabular}
}
\label{example}
\end{table*}

\section{Related Work}
\label{sec:Related Work}

 A significant body of work has established that speech and language pattern changes correlate with cognitive decline. Recent advances in natural language processing (NLP), and machine learning (ML) promise to assist clinicians in diagnosing Alzheimer’s disease and other types of dementia.
 Orimaye et al. \cite{orimaye2014learning} extracted the lexical and syntactic features from the DementiaBank Pitt corpus for the binary distinction between dementia patients from healthy people with an F1-measure of 74\%. Rentoumi et al. \cite{rentoumi2014features} used a Naive Bayes classifier with similar features in identifying AD with and without vascular pathology.
 While Fraser et al. used both linguistic and acoustic features and achieved an accuracy of over 81\% in distinguishing individuals with AD (combining possible AD and probable AD) from those without AD (elderly controls) \cite{fraser2016linguistic}, we treat samples of possible AD and probable AD separately. 
 The most relevant work to ours is  Masrani et al. \cite{masrani2017domain} who demonstrated the viability of domain adaptation using DementiaBank MCI data in a binary-classification task. They predicted MCI while training on MCI and AD samples, where the target data (class 1) contained 86 samples (43 MCI, 41 control), and the source data (class 2) contained 458 samples (236 probable AD, 21 possible AD, 201 control). To emphasize, they split the interviews from elderly controls to merge with their two classes of target (MCI) and the source (probable/possible AD) interviews. We treat each diagnosis group separately in a multi-class setting instead of splitting elderly controls as part of either target (MCI) or source (AD) groups. In a large population reflecting a range of cognitive decline, it is important to distinguish MCI from people with possible and probable AD as it is to distinguish them from elderly controls.

\section{Method}
\label{sec:Methodology}

We trained our classification models using a 10-fold cross-validation procedure. 
We used stratified sampling to ensure that the dataset's training and test split have the same distribution for each class. Before training the models, we extracted 108 theory-driven psycholinguistic features for all the transcripts from the Coh-Metrix computational tool. We first normalized the features and then calculated their variance. We removed the features with low variance (with a threshold of 0.0001) because it does not improve the model performance. Finally, we retained 88 features. In order to minimize overfitting and improve generalization in the face of a highly imbalanced dataset with the smallest class having only 21 samples (possible AD), we applied Lasso Regression (L1 regularization)\cite{tibshirani1996regression} and retained the 28 most dominant features. 
\begin{figure}[htbp] %!t
\centering
\frame{\includegraphics[width=0.75\columnwidth]{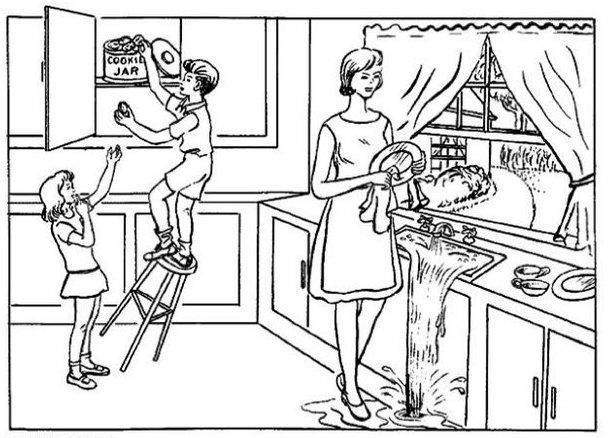}}
\caption{The Cookie Theft picture. Participants were asked to describe what they see in the image.}
\label{cookie}
\end{figure}
First we pre-processed all  transcripts by converting to lower case, removing all special characters and performing contraction mapping. Then, we used pre-trained BERT uncased model weights for the data-driven contextual language representations to generate 1024 dimension embeddings for each transcript. To avoid overfitting, we used Principal Component Analysis (PCA) on the embeddings to reduce the dimensionality of the feature space to 300. Due to high-class imbalance, we utilized the Synthetic Minority Oversampling Technique (SMOTE)\cite{chawla2002smote} only on training data features and embeddings. To identify the diagnosis stage and type of cognitive decline from verbal utterances, we formulate a multi-class classification problem with four classes (healthy control, MCI, possible AD and probable AD).

\subsection{Pitt Corpus: DementiaBank}
\label{sec:data}
We used the Pitt Corpus, a resource from the DementiaBank repository \cite{becker1994natural} that has been frequently used by computational linguists. 
It consists of 550 transcripts and recordings of 310 English-speaking participants including 209 people with some stage and type of dementia and 102 normal elderly controls (aged between 45-90 years), describing the ``Cookie Theft Picture" (Figure \ref{cookie}).  We used 242, 43, 21, and 236 transcripts belonging to participants clinically diagnosed to be elderly control, people with MCI, possible AD and probable AD, respectively. 
The remaining interviews were not used in this study due to significantly low sample numbers (Vascular AD : 3, Memory AD : 3, Other : 1). Table \ref{example} shows example responses from participants diagnosed under the four diagnostic categories considered in this study. We correctly identified all of these transcripts. 

\subsection{Classification Features}
\label{data_driven}
Feature vector representation plays a crucial role. While traditional context-free representations such as Bag-of-Words (BoW) or TF-IDF are conventional approaches to word-level feature representation, recent work incorporates more sophisticated language models.
We compare the results of our classification experiments with contextual BERT embeddings and psycholinguistic Coh-Metrix features as discussed in Section \ref{sec:Experiments and Results}. We do not report the results of our classification experiments with other BoW or TF-IDF features as they performed poorly in identifying MCI and possible AD. We designed our models with Linear Regression (LR), Logistic Regression (LogR), Multinomial Na\"ive Bayes (MNB) and Support Vector Machines (SVM) algorithms. Again, we do not report the results with LR and MNB as they failed to identify the samples of people with MCI and possible AD. We report the results with SVM across different classification features in multiple settings. In this study we used two types of features:
\begin{enumerate}
    \item Language Models (Contextual Embedding): We utilize BERT \cite{devlin2018bert} embeddings to learn the distributional contextual representation of words, representing the context. 
    \item Psycholinguistic features: As discourse is recognized as a fundamental component in language assessments, we extracted 108 theory-driven psycholinguistic features (indices) for each transcript from the Coh-Metrix computational tool\footnote{http://cohmetrix.memphis.edu/cohmetrixhome/documentation\_indices.html}. 
\end{enumerate}

\section{Results and Discussion}
\label{sec:Experiments and Results}

We conducted two sets of experiments. The first set of experiments tested for an early indicator of dementia; i.e., MCI from different stages in gradual cognitive decline (elderly controls, MCI, probable AD) (Table \ref{res:classification}). The second set of experiments tested for the various types of cognitive decline: MCI, sudden onset of dementia i.e., possible AD and a more severe form of dementia i.e., probable AD) (Table \ref{res:classification_4class}). 
We evaluated model performance with three different input feature groupings: (1) Theory-driven numerical psycholinguistic features extracted from Coh-Metrix, (2) Data-driven contextual BERT embeddings, and (3) Concatenation of both Coh-Metrix features and BERT embeddings. In (3), we investigated the viability of combining both theory-driven and data-driven features. 
We report the micro-averaged metrics as it treats each class equally when there is a class imbalance. Accuracy can be sensitive to class imbalance, whereas precision, recall, and micro-averaged F1-measures are asymmetric metrics. 

\begin{table*}
\centering
\caption{Multi-class Classification Results for Elderly Controls, Mild Cognitive Impairment (MCI) and Probable AD (AD:Alzheimer's Disease) speech transcripts}
\scalebox{1.0}{
\vline
\begin{tabular}{c|c|ccc|ccc|ccc|}
\cline{1-11}

\multicolumn{1}{c}{Model}
\vline
&\multicolumn{1}{c}{Dimension}
\vline
& \multicolumn{3}{c}{Elderly Controls }
\vline
& \multicolumn{3}{c}{MCI }
\vline
& \multicolumn{3}{c}{Probable AD }
\vline\\
\multicolumn{1}{c}{}
\vline
&\multicolumn{1}{c}{}
\vline
& \multicolumn{1}{c}{Micro-F1}
& \multicolumn{1}{c}{Precision}
& \multicolumn{1}{c}{Recall}
\vline
& \multicolumn{1}{c}{Micro-F1}
& \multicolumn{1}{c}{Precision}
& \multicolumn{1}{c}{Recall}
\vline
& \multicolumn{1}{c}{Micro-F1}
& \multicolumn{1}{c}{Precision}
& \multicolumn{1}{c}{Recall}
\vline\\ \cline{1-11}
Coh-Metrix(SVM)             &28 & 0.46 &0.57 &0.40 &0.06 &0.05 &0.13 &0.56 &0.56 &0.57 \\
BERT(SVM)                   &300 &0.81 &0.81 &0.79 &0.13 &0.14 &0.13 &0.76 &0.73 &0.78 \\
\textbf{Coh-Metrix + BERT (SVM)}      &328 &0.75 &0.73 &0.77 &0.37 &0.37 &0.37 &0.77 &0.79 &0.73 \\
\cline{1-11}
\end{tabular}
}
\label{res:classification}

\vspace{0.6cm}

\centering
\caption{Multi-class Classification Results for Elderly Controls, Mild Cognitive Impairment (MCI), Possible AD and Probable AD (AD:Alzheimer's Disease) speech transcripts}
\scalebox{0.82}{
\vline
\begin{tabular}{c|c|ccc|ccc|ccc|ccc|}
\cline{1-14}
\multicolumn{1}{c}{Algorithm}
\vline
&\multicolumn{1}{c}{Dimension}
\vline
& \multicolumn{3}{c}{Elderly Control }
\vline
& \multicolumn{3}{c}{MCI }
\vline
& \multicolumn{3}{c}{Possible AD}
\vline
& \multicolumn{3}{c}{Probable AD}
\vline\\
\multicolumn{1}{c}{}
\vline
&\multicolumn{1}{c}{}
\vline
& \multicolumn{1}{c}{Micro-F1}
& \multicolumn{1}{c}{Precision}
& \multicolumn{1}{c}{Recall}
\vline
& \multicolumn{1}{c}{Micro-F1}
& \multicolumn{1}{c}{Precision}
& \multicolumn{1}{c}{Recall}
\vline
& \multicolumn{1}{c}{Micro-F1}
& \multicolumn{1}{c}{Precision}
& \multicolumn{1}{c}{Recall}
\vline
& \multicolumn{1}{c}{Micro-F1}
& \multicolumn{1}{c}{Precision}
& \multicolumn{1}{c}{Recall}\vline\\ \cline{1-14}
Coh-Metrix(SVM)             &28 &0.44 &0.53 &0.38 &0.07 &0.05 &0.13 &0.33 &0.33 &0.33 &0.42 &0.53 &0.35\\
BERT(SVM)                   &300 &0.80 &0.81 &0.79 &0.14 &0.17 &0.13 &0.22 &0.20 &0.25 &0.70 &0.69 &0.72 \\
\textbf{Coh-Metrix + BERT (SVM)}      &328 &0.74 &0.74 &0.75 &\textbf{0.28} &0.28 &0.28 &\textbf{0.35} &\textbf{0.48} &\textbf{0.67} &\textbf{0.73} &\textbf{0.72} &\textbf{0.73} \\
\cline{1-14}
\end{tabular}
}
\label{res:classification_4class}
\end{table*}

The results with other BoW, Word2Vec and TF-IDF models were poor in identifying elderly controls and probable AD. Hence, we only report the results with the features that were able to distinguish between all diagnosis groups.
First, we designed a binary classifier to separate AD (possible AD and probable AD) vs. non-AD (elderly controls) and observed that our SVM model trained on 328 dimension vector was able to identify correctly 81.3\% of the samples as compared to Fraser et al.\cite{fraser2016linguistic} (81\% accuracy with 370 linguistic and acoustic features). This shows the significance of our less complex models designed on contextual and psycholinguistic features to identify AD with high accuracy.

Second, we formulate a three-class classification problem (i.e., transcripts of healthy elderly controls, people with MCI and probable AD) to identify the early indicators of gradual cognitive decline (MCI). To understand our model's feasibility to identify a new verbal utterance from a person diagnosed with possible AD, we used the best performing model (Table \ref{res:classification} row 3) to predict labels on the transcripts from people with possible AD. More than 50\% (13 out of 21) samples were identified as probable AD, indicating that the model had learned the similarity between verbal utterances of people with possible AD and probable AD, irrespective of their origin/cause. To distinguish the samples of people diagnosed with probable AD and possible AD, we designed a four-class classification setting (elderly healthy controls, people with MCI, possible AD and probable AD). 

In all of our experiments, our model trained with contextual embeddings was able to identify more verbal utterances with elderly healthy controls correctly, given that healthy people display a sound contextual memory. Our models were able to identify more transcripts of people with MCI and probable AD correctly with new psycholinguistic features as they captured the loss of coherency - a characteristic of cognitive decline. The low precision and recall in MCI and possible AD is indicative of the challenge in distinguishing them from the speech patterns of elderly controls and probable AD respectively with highly imbalanced and limited data. 

We believe that the decline in context memory and the variability in linguistic characteristics complement each other to effectively distinguish the spoken language patterns to identify the various stages and types of cognitive decline. Using contextual language embeddings, the combined model trained on both data and theory-driven features was able to capture the contextual memory deficits. This is consistent with the fact that elderly controls display better context memory in verbal utterances than people with cognitive impairment who tend to lose the thread of a conversation. 
For example, as shown in Table \ref{example}, highlighted in blue, the person with MCI tends to go out of context (Figure \ref{cookie}) and starts talking about how he/she uses a dishwasher to do the dishes at home, which is just pots and pans, and that they consume eggs in the morning. The person with probable AD loses context in the middle, talking about how the father is not coming while there is no indication of a fatherly image in the picture.

\section{Conclusion}
\label{conclusion}

Identifying the symptoms of cognitive decline at an early stage can help  clinicians prescribe therapies and encourage lifestyle and dietary changes, which can be helpful for managing the onset of dementia. We used state-of-the-art contextual language models and combined them with psycholinguistic  macro-level features to understand the language dysfunction and decline in context memory with cognitive impairment. Our results show that machine learning models trained on both psycholinguistic and contextual models can distinguish early indicators of cognitive decline (MCI) as well as multiple types of AD (possible AD and probable AD) from verbal utterances of elderly controls in a multi-class setting. 
We plan to extend this work to automatically identify various speech symptoms, such as repetitiveness, confusion, problems with thinking and reasoning, and forgetting the names of everyday objects. This method could be particularly useful in developing countries, which have an even more pressing need for inexpensive and non-invasive solutions.

\section{Acknowledgment}
The authors are deeply grateful to the SoonJye Kho and Amanuel Alambo for peer-reviewing the paper.
\bibliographystyle{IEEEtran}
\bibliography{IEEEabrv,sample.bib}
\end{document}